\documentclass[a4paper,twoside]{article}

\usepackage{epsfig}
\usepackage{subcaption}
\usepackage{booktabs}
\usepackage{calc}
\usepackage{amssymb}
\usepackage{amstext}
\usepackage{amsmath}
\usepackage{amsthm}
\usepackage{multicol}
\usepackage{pslatex}
\usepackage{apalike}
\usepackage{algorithm2e}
\usepackage[bottom]{footmisc}
\usepackage{balance}
\usepackage{url}
\usepackage{SCITEPRESS}     
\usepackage{url} 

\begin{document}

\title{Spatio-Temporal Transformers for Long-Term NDVI Forecasting}

\author{\authorname{Ido Faran\sup{1}, Nathan S. Netanyahu\sup{1,2}, and Maxim Shoshany\sup{3}}
\affiliation{\sup{1}Dept. of Computer Science, Bar-Ilan University, Ramat Gan 5290002, Israel}
\affiliation{\sup{2}Dept. of Computer Science, College of Law and Business, Ramat Gan 5257346 , Israel}
\affiliation{\sup{3}Faculty of Civil and Environmental Engineering, Technion Israel Institute of Technology, Haifa 3200003, Israel}
}


\abstract{
Long-term satellite image time series (SITS) analysis in heterogeneous landscapes faces significant challenges, particularly in Mediterranean regions where complex spatial patterns, seasonal variations, and multi-decade environmental changes interact across different scales. This paper presents the Spatio-Temporal Transformer for Long Term Forecasting (STT-LTF
), an extended framework that advances beyond purely temporal analysis to integrate spatial context modeling with temporal sequence prediction. STT-LTF processes multi-scale spatial patches alongside temporal sequences (up to 20 years) through a unified transformer architecture, capturing both local neighborhood relationships and regional climate influences. The framework employs comprehensive self-supervised learning with spatial masking, temporal masking, and horizon sampling strategies, enabling robust model training from 40 years of unlabeled Landsat imagery. Unlike autoregressive approaches, STT-LTF directly predicts arbitrary future time points without error accumulation, incorporating spatial patch embeddings, cyclical temporal encoding, and geographic coordinates to learn complex dependencies across heterogeneous Mediterranean ecosystems. Experimental evaluation on Landsat data (1984-2024) demonstrates that STT-LTF achieves a Mean Absolute Error (MAE) of 0.0328 and R² of 0.8412 for next-year predictions, outperforming traditional statistical methods, CNN-based approaches, LSTM networks, and standard transformers. The framework's ability to handle irregular temporal sampling and variable prediction horizons makes it particularly suitable for analysis of heterogeneous landscapes experiencing rapid ecological transitions.
}

\onecolumn \maketitle \normalsize \setcounter{footnote}{0} \vfill
\section{\uppercase{Introduction}}

Forecasting future satellite imagery is a key challenge in computer vision and remote sensing \cite{zhu2017,kennedy2018}. Time-series forecasting aims to reveal temporal patterns that are often hidden in spatially complex landscapes where short-term, medium-term, and long-term processes occur and interact simultaneously across different spatial scales \cite{verbesselt2010detecting}. This task is particularly difficult for heterogeneous landscapes because daily, seasonal, and multi-decade changes overlap, making it hard to forecast the next state from past observations \cite{gao2022earthformer}. The complexity stems from the interplay of ecological, climatic, and human-driven factors, each contributing in distinct ways to the observed environmental conditions \cite{zhang2023crossformer}.

Effectively addressing these challenges requires models capable of extracting and integrating both temporal sequences and spatial contexts inherent in environmental data \cite{liu2023itransformer}. Traditional approaches often fall short due to their limited capacity to capture these multi-dimensional interactions, particularly when faced with extended temporal horizons and heterogeneous landscapes \cite{gomez2016optical}. Recent advancements in transformer-based architectures \cite{vaswani2017attention} have introduced powerful tools for modeling such complex dependencies by leveraging attention mechanisms capable of learning intricate relationships within and between spatial and temporal dimensions.

Self-supervised machine-learning techniques provide a significant advantage for addressing these challenges, as they operate without explicit labels and assumptions, enabling models to autonomously learn intrinsic representations from historical data. By training on extensive unlabeled satellite-image archives, these approaches inherently capture meaningful temporal and spatial relationships crucial for predicting future environmental conditions.

Earth-observation satellites have become invaluable tools for analyzing dynamic environmental processes, collecting data on our planet's surface and ecosystems for over half a century. The Landsat mission, operational since 1984, captures multispectral imagery globally every 16 days at a 30[m] spatial resolution. The resulting Satellite Image Time Series (SITS) adds a temporal dimension to Earth-observation data, supporting long-term monitoring of vegetation health, land-use change, and ecosystem dynamics \cite{zhu2019}. These capabilities are particularly crucial in the Mediterranean basin, especially its southeastern region, which has been recognized as a climate-change hotspot since the 1980s. This region faces increasing frequencies of droughts, forest and shrubland fires, agricultural abandonment, and progressive urbanization - all expected to intensify over the next two decades. Long-term satellite imagery time series provide essential data for understanding and predicting these complex environmental transformations.

\begin{figure*}[ht]
    \centering
    \begin{subfigure}[b]{0.48\textwidth}
        \centering
        \includegraphics[width=\textwidth, height=5cm, keepaspectratio]{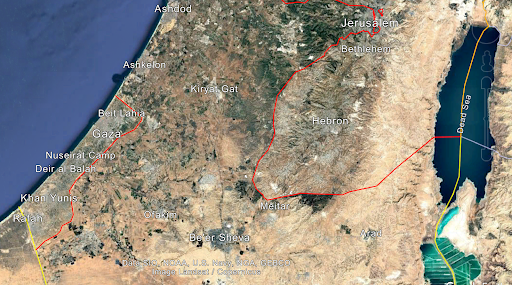}
        \caption{Study area (Google Earth)}
    \end{subfigure}
    \hfill
    \begin{subfigure}[b]{0.48\textwidth}
        \centering
        \includegraphics[width=\textwidth, height=5cm, keepaspectratio]{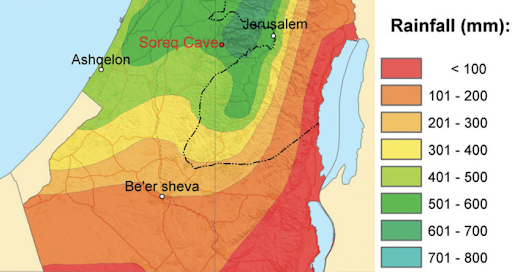}
        \caption{Rainfall distribution}
    \end{subfigure}
    
    \vspace{0.5cm} 
    
    \begin{subfigure}[b]{0.6\textwidth}
        \centering
        \includegraphics[width=\textwidth, height=6cm, keepaspectratio]{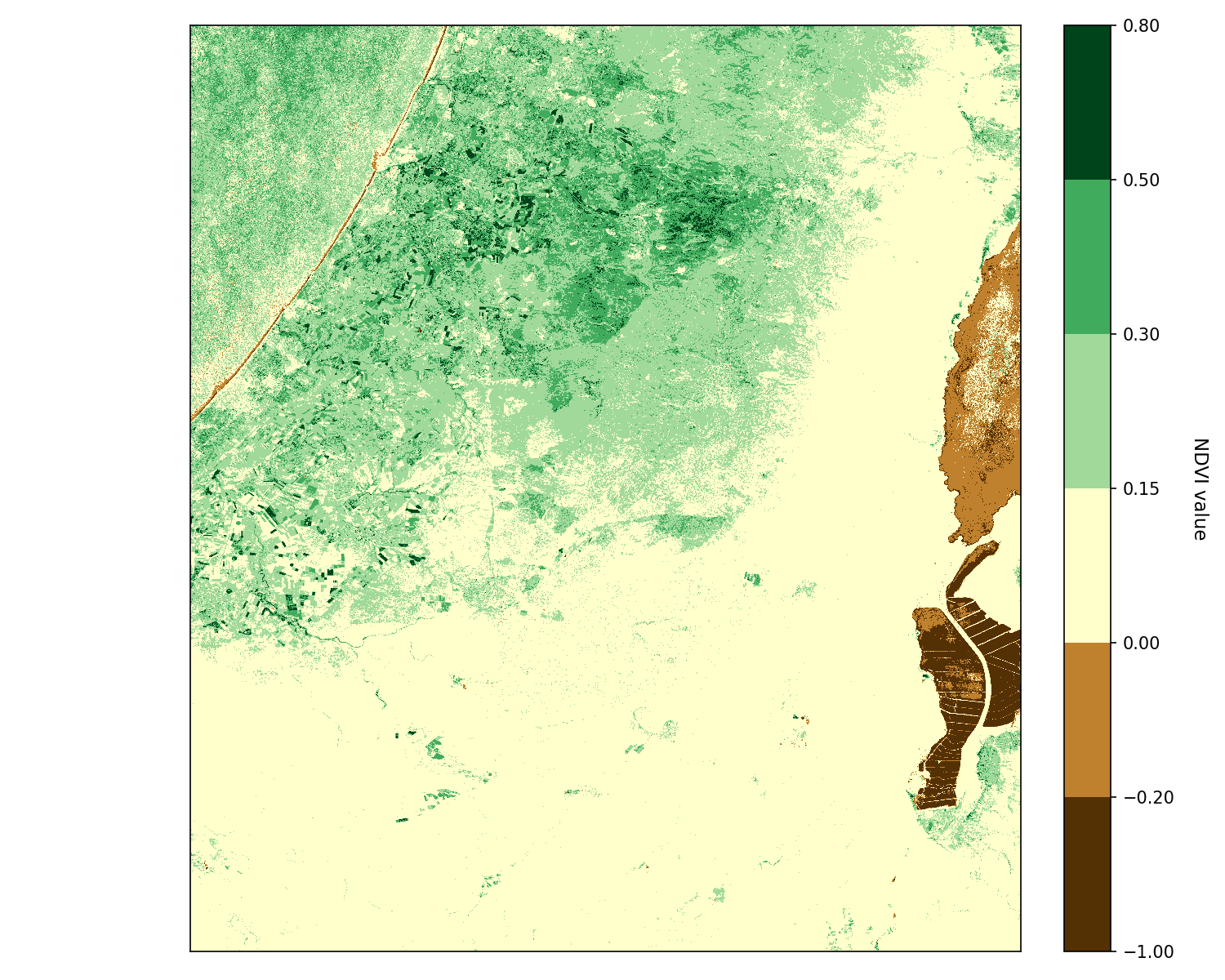}
        \caption{NDVI patterns}
    \end{subfigure}
    
    \vspace{0.5cm} 
    
    \caption{Study area characteristics in the southeastern Mediterranean basin.}
    \label{fig:study_area_combined}
\end{figure*}

\begin{figure*}[htbp]
\centerline{\includegraphics[width=2\columnwidth]{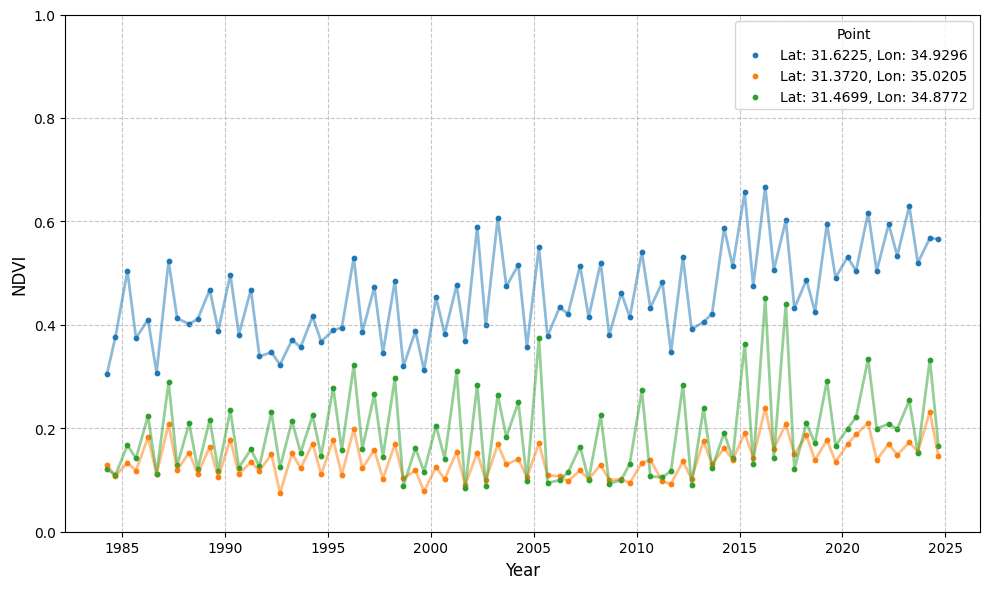}}
\caption{Time series of NDVI values (1985–2024) for three samples, illustrating seasonal variability and long-term trends. Source: \cite{faran2025sstltp}}
\label{fig:time_series_ndvi}
\end{figure*}

In this study, we utilize the Normalized Difference Vegetation Index (NDVI), a spectral vegetation index calculated as
\begin{equation}
\text{NDVI}=\frac{\text{Near Infrared Band} - \text{Red Band}}{\text{Near Infrared Band} + \text{Red Band}}
\end{equation}

which effectively represents photosynthetic activity, vegetation cover, and biomass. In Mediterranean climates, characterized by a short rainy season followed by an extended hot and dry period, NDVI values closely correlate with seasonal and annual rainfall variability (see Figure~\ref{fig:study_area_combined} and Figure \ref{fig:time_series_ndvi}). Predicting spatio-temporal NDVI patterns years and decades into the future is essential to understanding how ecosystems respond to climatic and anthropogenic pressures; a critical concern in these regions is desertification, particularly the stability or shift of desert boundaries due to climate change \cite{roitberg2024primary}.

One of the primary goals of time-series analysis in remote sensing is to accurately forecast future values based on historical observations. These capabilities support image prediction, gap filling caused by cloud cover or sensor issues, and integration of multiple data sources. However, the task faces unique challenges, including distinguishing climate-driven trends from natural variability, handling seasonality, and overcoming data gaps and inconsistencies \cite{zhu2017change}. Addressing these complexities is essential for robust environmental monitoring and predictive modeling over extended periods.

While existing approaches have made significant progress in satellite image time series analysis, a critical limitation persists in their capacity to fully integrate extended temporal patterns with spatial dependencies and location-specific characteristics. Traditional methods typically focus on temporal sequences spanning up to one year and often fail to capture the complex interplay between multiple dimensions over longer periods \cite{chen2024anthropogenic,li2021long}. This limitation becomes particularly pronounced in heterogeneous Mediterranean landscapes, where intricate seasonal patterns, anthropogenic disturbances, climatic variations, and location-dependent factors necessitate a comprehensive analytical framework operating across multiple spatial and temporal scales \cite{mehmood2024analyzing,xu2024anthropogenic}.

To address these challenges, we present an enhanced methodology called the Spatial Temporal Transformer for Long-Term Forecasting
(STT-LTF), a framework that significantly extends our previous work \cite{faran2025sstltp} by incorporating spatial context modeling alongside temporal sequence analysis. While our earlier approach focused exclusively on temporal prediction using pixel-wise NDVI time series, this extended framework introduces a fundamental paradigm shift from purely temporal modeling to integrated spatio-temporal representation learning \cite{cong2022satmae,noman2024rethinking}. The enhanced architecture incorporates multi-scale spatial patch embeddings that capture local neighborhood relationships and broader landscape patterns, enabling the model to understand not only when changes occur temporally but also how they propagate spatially across heterogeneous Mediterranean landscapes \cite{yang2025enhanced}. This spatial-temporal integration addresses limitations of pixel-independent approaches by leveraging spatial correlations to improve prediction accuracy, particularly for regions with sparse temporal observations or irregular sampling patterns \cite{li2024spatial}.

STT-LTF fuses spatial patch processing with temporal transformer architectures. Rather than treating pixels independently or separating spatial and temporal analyses, our framework processes N×N pixel patches throughout entire temporal sequences, capturing spatial relationships from single pixels up to 270[m] × 270[m] neighborhoods. We achieve this through self-supervised learning with extensive masking techniques, enabling the model to discover complex spatio-temporal patterns in unlabeled satellite data spanning multiple decades without manual annotation.

A key distinguishing feature of our approach is the direct prediction capability for arbitrary future time points. While conventional methods rely on recursive predictions that accumulate errors over extended horizons, STT-LTF learns to map directly from historical spatio-temporal patterns to future vegetation states at any specified target time. This is achieved through a novel temporal embedding strategy that treats the target time as an input parameter, allowing the model to learn the complex non-linear relationships between past observations and future conditions across varying temporal distances. The transformer architecture, specifically optimized for remote sensing data, employs specialized attention mechanisms that capture both local vegetation dynamics and regional climate influences, providing a more comprehensive understanding of ecosystem evolution.

Our framework addresses several fundamental challenges in long-term vegetation monitoring. First, it handles the inherent multi-scale nature of vegetation dynamics, where local disturbances interact with regional climate patterns over different temporal scales. Second, it provides a unified approach for analyzing heterogeneous landscapes without requiring separate models for different vegetation types or geographical regions. Third, it offers computational efficiency by eliminating the need for iterative predictions while maintaining high accuracy across extended forecast horizons. These innovations make STT-LTF particularly suitable for applications requiring reliable long-term vegetation predictions, such as climate impact assessment, agricultural planning, and ecosystem management in Mediterranean and similar complex environments.

The main contributions of this work are as follows:

\begin{enumerate}
\item \textbf{Advanced spatio-temporal embedding framework with direct prediction capability}: We introduce a versatile embedding architecture that seamlessly integrates spatial patches (1×1 to 9×9 pixels), temporal sequences (2.5 to 20 years), and geographical location information through continuous representation spaces. The framework captures both periodic seasonal patterns and long-term directional trends, enabling direct prediction at any future time point without recursive error accumulation. This unified approach supports predictions across arbitrary spatial and temporal resolutions while maintaining accuracy over extended forecast horizons.

\item \textbf{Self-supervised learning from unlabeled multi-decade data}: We present a novel self-supervised training method specifically designed for long-term SITS analysis, capable of learning complex vegetation dynamics from 40 years of unlabeled Landsat imagery without requiring manual annotations. This approach leverages the temporal structure inherent in satellite data with comprehensive masking strategies to create a powerful predictive model from incomplete time series.

\item \textbf{Transformer architecture for long-term spatio-temporal prediction}: We design a specialized transformer that processes extended temporal sequences with spatial context through multi-scale attention mechanisms optimized for vegetation dynamics. The architecture features adaptive encodings for irregular satellite sampling patterns and non-autoregressive prediction heads that directly forecast future vegetation states at arbitrary time horizons, avoiding the error accumulation inherent in recursive approaches.

\item \textbf{Comprehensive evaluation on Mediterranean ecosystems}: We conduct systematic experiments on Landsat NDVI data from heterogeneous Mediterranean landscapes, analyzing the individual and combined impacts of spatial resolution, temporal sequence length, and forecast horizon on prediction accuracy. Our evaluation demonstrates the framework's adaptability across diverse vegetation types and environmental conditions, achieving state-of-the-art performance in both short-term and long-term predictions.
\end{enumerate}

\begin{figure*}[t]
\centerline{\includegraphics[width=2\columnwidth]{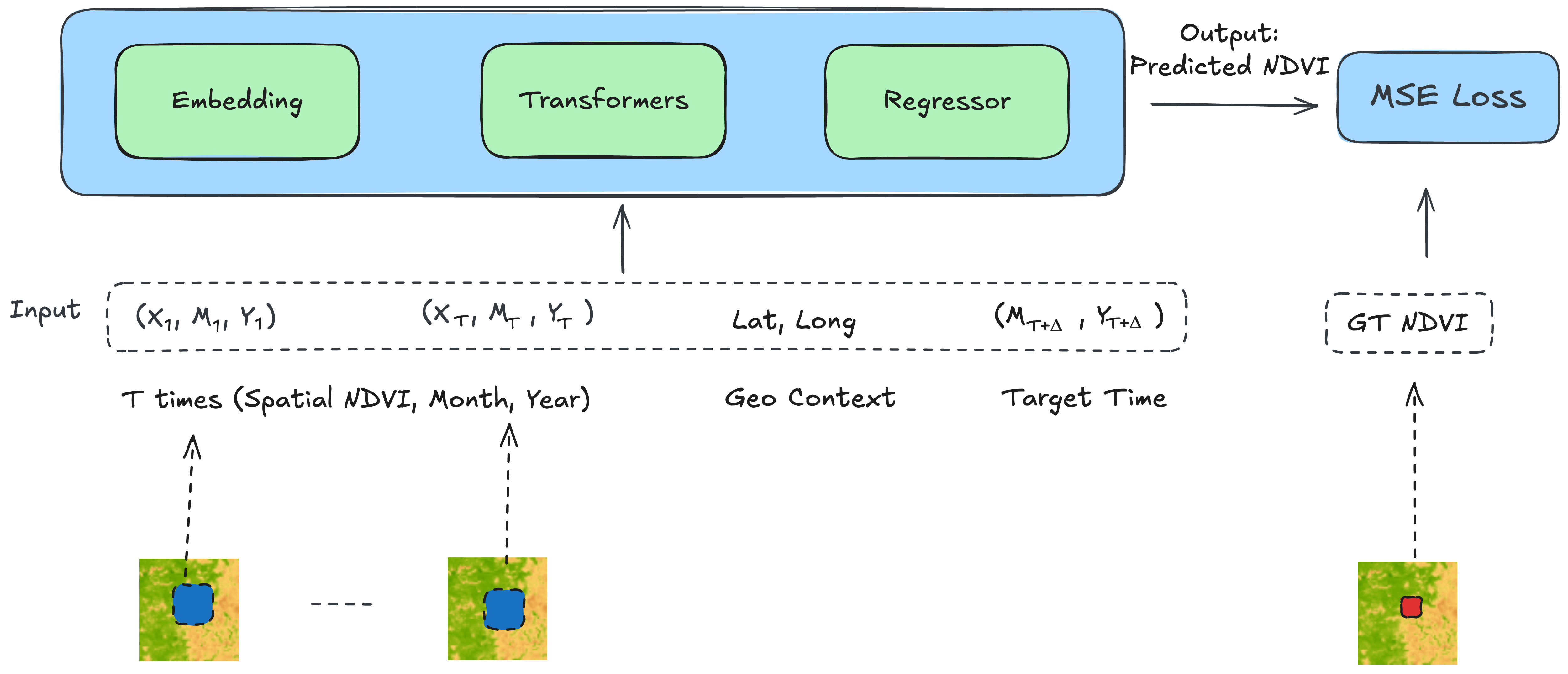}}
\caption{Self-supervised learning framework for NDVI time-series prediction. The model processes $T$ historical satellite observations with spatial patches, each containing NDVI, month, and year and geographic context to predict future NDVI values at target time $T+\Delta$ using MAE loss against ground truth NDVI from actual satellite observations.}
\label{fig:training_method}
\end{figure*}

The remainder of this paper is organized as follows: Section 2 reviews related work covering traditional statistical methods, deep learning approaches, and self-supervised learning techniques for time-series analysis. Section 3 presents the proposed methodology, detailing the input sampling and masking strategy, spatio-temporal embedding framework, transformer encoder architecture, and regression decoder. Section 4 describes the experimental setup and evaluates model performance across different spatial resolutions and temporal horizons using Mediterranean Landsat data. Section 5 concludes with a summary of findings and future research directions.

\section{\uppercase{Related Work}}
\subsection{Classical Statistical Approaches in Remote Sensing Time Series}
Classical statistical approaches have formed the backbone of remote sensing time series analysis, employing methods such as Random Forests, Cellular Automata Markov Chain models, and Autoregressive Integrated Moving Average (ARIMA) frameworks \cite{gomez2016optical}. Seasonal patterns are captured through Seasonal ARIMA (SARIMA) and Facebook Prophet models \cite{box2015time,taylor2018forecasting}, which excel at handling cyclical behaviors in vegetation monitoring applications.

Recent applications demonstrate the continued effectiveness of ARIMA-based methods. \cite{tian2016drought} successfully applied ARIMA models for drought forecasting using satellite-derived Vegetation Temperature Condition Index data, while \cite{spatial2024forecasting} achieved high accuracy in solar radiation prediction using SARIMA models with NASA's satellite data archive. These studies highlight the robustness of statistical methods for multi-decade environmental datasets.

Hybrid approaches combining statistical foundations with machine learning have emerged as promising alternatives. SARIMA-Neural Network combinations leverage seasonal pattern recognition while incorporating non-linear learning capabilities \cite{ruiz2014hybrid}. Recent comparative studies \cite{forecasting2024energy} revealed that SARIMA models consistently outperformed Support Vector Regression for energy forecasting when seasonal patterns were present.

Notably, \cite{analysis2024effectiveness} found that SARIMA models outperformed advanced deep learning approaches, including Long Short-Term Memory networks for wind speed forecasting, suggesting that statistical methods retain advantages in interpretability and computational efficiency for operational applications.

However, traditional statistical approaches face limitations with complex non-linear relationships and irregular temporal patterns in heterogeneous landscapes. Missing data handling remains challenging, particularly with cloud-affected satellite imagery, motivating the development of more sophisticated spatio-temporal modeling approaches.

\subsection{Deep Learning Architectures in Remote Sensing}

The evolution of deep learning architectures has fundamentally transformed how researchers approach remote sensing time-series analysis. Early innovations with Convolutional Neural Networks demonstrated the potential for automatic feature extraction from satellite imagery, eliminating the need for handcrafted features \cite{zhu2017}. Long Short-Term Memory networks subsequently proved effective in capturing temporal dependencies within satellite time-series, particularly for applications like crop classification and phenological monitoring where seasonal patterns are crucial.

Contemporary deep learning frameworks for remote sensing have evolved to address domain-specific challenges through specialized architectural designs. The integration of spatial and channel attention mechanisms allows networks to focus on relevant spectral bands and spatial regions, improving discrimination between similar land cover types \cite{applicationdeeplearning2023}. Multi-scale feature extraction approaches, inspired by the varying ground sampling distances in satellite imagery, enable models to simultaneously process objects of different sizes within a single framework.

Recurrent architectures have been particularly successful in modeling temporal dynamics of Earth observation data. Bidirectional LSTMs capture both forward and backward temporal dependencies, proving valuable for detecting abrupt changes and gradual transitions in land cover. Gated Recurrent Units (GRUs) offer computational efficiency while maintaining temporal modeling capabilities, making them suitable for processing extensive satellite archives. The combination of convolutional and recurrent layers in hybrid architectures enables simultaneous extraction of spatial features and temporal patterns, addressing the inherent spatio-temporal nature of remote sensing data \cite{remotesensingreview2024}. These architectures have shown remarkable success in applications ranging from vegetation dynamics monitoring to urban growth prediction.

\subsection{Transformer-based Approaches for Temporal Modeling}

The adaptation of Transformer architectures to remote sensing has catalyzed a paradigm shift in processing satellite time-series data. Unlike their application in natural language processing, Transformers in remote sensing must handle irregular temporal sampling, missing data due to cloud coverage, and multi-modal sensor inputs \cite{transformersreview2024}. Recent innovations have focused on developing specialized attention mechanisms that accommodate these domain-specific challenges.

State-space models represent an emerging direction for efficient temporal modeling in remote sensing. The Mamba architecture and its variants have shown promise for processing long sequences with linear computational complexity \cite{visionmamba2025}. While primarily explored for change detection and segmentation tasks, these models offer potential advantages for time-series analysis through their selective state-space mechanisms that can adaptively focus on relevant temporal features \cite{changemamba2024}.

Multi-scale temporal attention mechanisms have proven essential for capturing phenomena occurring at different temporal frequencies. The Spatio-Temporal SwinMAE architecture introduces hierarchical temporal modeling that processes seasonal patterns, inter-annual variations, and rapid changes simultaneously \cite{githubremotesensing2024}. This multi-scale approach enables accurate monitoring of diverse Earth processes, from gradual land degradation to sudden natural disasters.

Recent benchmarking studies reveal that lightweight Transformer variants specifically designed for remote sensing outperform generic architectures \cite{transformermodels2024}. These models incorporate domain-specific inductive biases, such as temporal position encodings based on acquisition dates and sensor-specific normalization schemes, resulting in improved generalization across different geographic regions and temporal periods.

\subsection{Self-Supervised Learning Approaches}
Self-supervised learning has revolutionized remote sensing by addressing the fundamental challenge of acquiring labeled data across vast geographical regions. The annotation of satellite imagery demands specialized expertise and substantial resources, particularly when dealing with diverse ecosystems and temporal variations \cite{wang2022selfsupervised}. Recent developments have demonstrated that self-supervised methods can leverage the abundant unlabeled satellite data to learn meaningful representations without manual annotations.

Modern self-supervised frameworks for remote sensing employ various pretext tasks tailored to the unique characteristics of Earth observation data. Contrastive learning approaches have shown remarkable success by learning invariant representations across different augmentations and viewing conditions \cite{selfsupreview2022}. The SimSiam framework, when pre-trained on high-resolution remote sensing datasets, has demonstrated superior transfer learning capabilities compared to ImageNet pre-trained models, particularly for land cover classification tasks \cite{indomain2024}. This domain-specific pre-training captures spectral and spatial patterns unique to satellite imagery that general-purpose models often miss.

Several frameworks have been developed to handle the temporal aspects of satellite data within self-supervised paradigms. The Presto model represents progress in handling pixel-time-series through lightweight transformer architectures, demonstrating effectiveness for agricultural monitoring applications \cite{tseng2023lightweight}. Temporal contrastive learning methods have been adapted to learn season-invariant features, enabling models to distinguish between natural phenological cycles and actual land cover changes \cite{yuan2020}. Multi-modal self-supervised approaches have also emerged, jointly learning from optical and radar data to create weather-independent representations \cite{selfsupreview2022}.

\subsection{Challenges and Gaps in Long-term Spatio-Temporal Analysis}

Despite significant advances in deep learning and self-supervised approaches, several critical challenges remain specific to long-term spatio-temporal modeling in remote sensing. Current self-supervised methods predominantly focus on single-image representations or short temporal sequences, typically spanning one year or less \cite{yuan2020}. Agricultural applications have received disproportionate attention, leaving complex heterogeneous landscapes underexplored \cite{russwurm2018}. This temporal limitation prevents models from capturing multi-decadal environmental processes essential for climate monitoring and land use change detection.

The computational demands of autoregressive prediction methods present significant obstacles for long-term forecasting. These approaches require sequential model activations for each temporal step, leading to error propagation and exponential growth in computational complexity \cite{miller2024}. Each prediction depends on previous outputs, causing small errors to compound over extended horizons. This cascading effect becomes particularly problematic when analyzing multi-decadal satellite archives where subtle long-term trends must be distinguished from noise and seasonal variations.

Data heterogeneity poses another fundamental challenge for long-term analysis. Satellite missions evolve over time, introducing sensor variations, orbital changes, and calibration differences that complicate multi-decadal studies. Existing models often fail to account for these systematic changes, resulting in temporal inconsistencies and spurious trends \cite{foundationmodelssurvey2024}. The lack of standardized preprocessing pipelines for harmonizing long-term satellite archives further exacerbates this issue.

Memory constraints represent a significant bottleneck for processing extensive temporal sequences. While state-space models offer linear complexity, they still require substantial memory for maintaining hidden states across long sequences. This limitation forces researchers to adopt temporal windowing strategies that may miss important long-term dependencies and cyclical patterns spanning multiple years. The trade-off between sequence length and model capacity remains a fundamental obstacle to capturing multi-decadal environmental dynamics.

The scarcity of labeled data for long-term changes presents a critical gap in model development and evaluation. Most benchmark datasets focus on short-term changes or single-time observations, failing to capture gradual environmental processes like desertification, urbanization, or climate-induced vegetation shifts \cite{remotesensingreview2024}. Additionally, the computational cost of training models on extensive temporal sequences often necessitates trade-offs between temporal resolution and sequence length, potentially missing critical transitional periods in environmental dynamics.

Addressing these limitations requires fundamental rethinking of self-supervised objectives for temporal remote sensing data. Rather than adapting computer vision methods designed for static images, the field needs approaches that inherently model temporal dynamics, seasonal cycles, and long-term environmental changes. Direct temporal pattern learning, as opposed to iterative prediction, offers promising directions for capturing extended-range dependencies while maintaining computational efficiency \cite{zhu2019}. Such methods must also account for the unique challenges of Mediterranean and other heterogeneous landscapes, where human activities, climate variability, and ecological processes interact across multiple temporal scales.

The integration of spatio-temporal transformers with self-supervised learning presents a particularly promising direction for addressing these challenges. By processing entire temporal sequences through unified architectures that combine spatial patches, temporal embeddings, and geographic context, models can learn complex dependencies without the cascading errors of autoregressive approaches. Self-supervised masking strategies that operate across both spatial and temporal dimensions enable models to learn from the inherent structure of satellite time series, leveraging the vast amounts of unlabeled data available. This approach is especially valuable for heterogeneous landscapes where labeled data spanning multiple decades is scarce. Direct prediction frameworks that forecast future values in a single forward pass, rather than through iterative steps, offer computational efficiency while maintaining the ability to capture long-range temporal dependencies. Such architectures can adapt to varying prediction horizons and spatial resolutions through flexible embedding schemes, making them suitable for diverse remote sensing applications from short-term vegetation monitoring to multi-decadal climate impact assessment.

\section{\uppercase{Proposed Method}}
Our proposed approach formulates time-series prediction of satellite NDVI as a self-supervised learning problem, designed to exploit both the temporal evolution and spatial context of long-term satellite observations. As illustrated in Figure~\ref{fig:training_method}, we feed the model sequences of past satellite imagery along with spatial patches centered around each target pixel, enabling it to learn from historical patterns without requiring labeled data. The training objective minimizes the difference between predicted and observed NDVI values, leveraging the rich spatio-temporal structure inherent in Landsat time series.

This methodology predicts the future NDVI value of a central pixel by considering both its temporal sequence and the surrounding spatial environment at each time step. This combination allows the model to capture local spatial patterns, seasonal cycles, and long-term trends simultaneously. Specifically, for every pixel of interest, we extract a sequence of past NDVI patches and encode them together with corresponding temporal information—such as year and month—and location coordinates. The model is trained to forecast the NDVI value at a future time point by comparing predictions against actual measurements and iteratively updating its parameters to reduce forecasting errors.

This spatio-temporal learning strategy enables the model to refine its predictions over time, adjusting its understanding of vegetation dynamics based on both historical observations and the spatial relationships of neighboring pixels. By integrating these dimensions, our method effectively learns complex dependencies that govern vegetation growth and response to climatic and anthropogenic factors across heterogeneous landscapes.

Formally, we represent the input as a time-series sequence that includes spatial, temporal, and geographic information:
\begin{equation}\label{eqinput}
O = \{(X_t, M_t, Y_t, \text{lat}, \text{lon}) \mid t = 0, \ldots, T\}
\end{equation}
where $T$ denotes the length of the historical time series, $X_t \in \mathbb{R}^{N \times N}$ represents the NDVI values within a spatial patch of size $N \times N$ at time $t$, $M_t$ indicates the month corresponding to time $t$, $Y_t$ specifies the year at time $t$, and $\text{lat}$ and $\text{lon}$ refer to the latitude and longitude of the patch’s central pixel.

The prediction target is defined by the future month and year, denoted as \( M_{T+\Delta} \) and \( Y_{T+\Delta} \), which specify the future timestamp for forecasting. The model’s task is to estimate the NDVI value at the central pixel for this future time point \( T + \Delta \), formulated as:
\begin{equation}
\hat{x}_{T+\Delta} = f(O, M_{T+\Delta}, Y_{T+\Delta})
\end{equation}
where \( f \) is the transformer-based deep learning model. Unlike autoregressive models that require sequential predictions for each intermediate step, our non-autoregressive approach allows direct forecasting of the NDVI value for any specified future timestamp in a single step. By incorporating the target month and year together with the past temporal sequence, spatial patch data, and geographic coordinates, the model avoids error accumulation over multiple iterations and enables efficient, flexible prediction across arbitrary future horizons.

The predicted value \( \hat{x}_{T+\Delta} \) is compared to the true observed value \( x_{T+\Delta} \) using a loss function \( L(\hat{x}_{T+\Delta}, x_{T+\Delta}) \). Model parameters are updated through iterative backpropagation to minimize this loss, thereby enhancing the model’s ability to forecast NDVI values accurately across different temporal horizons.

\begin{figure*}[htbp]
\centerline{\includegraphics[width=1\textwidth, height=10\textheight, keepaspectratio]{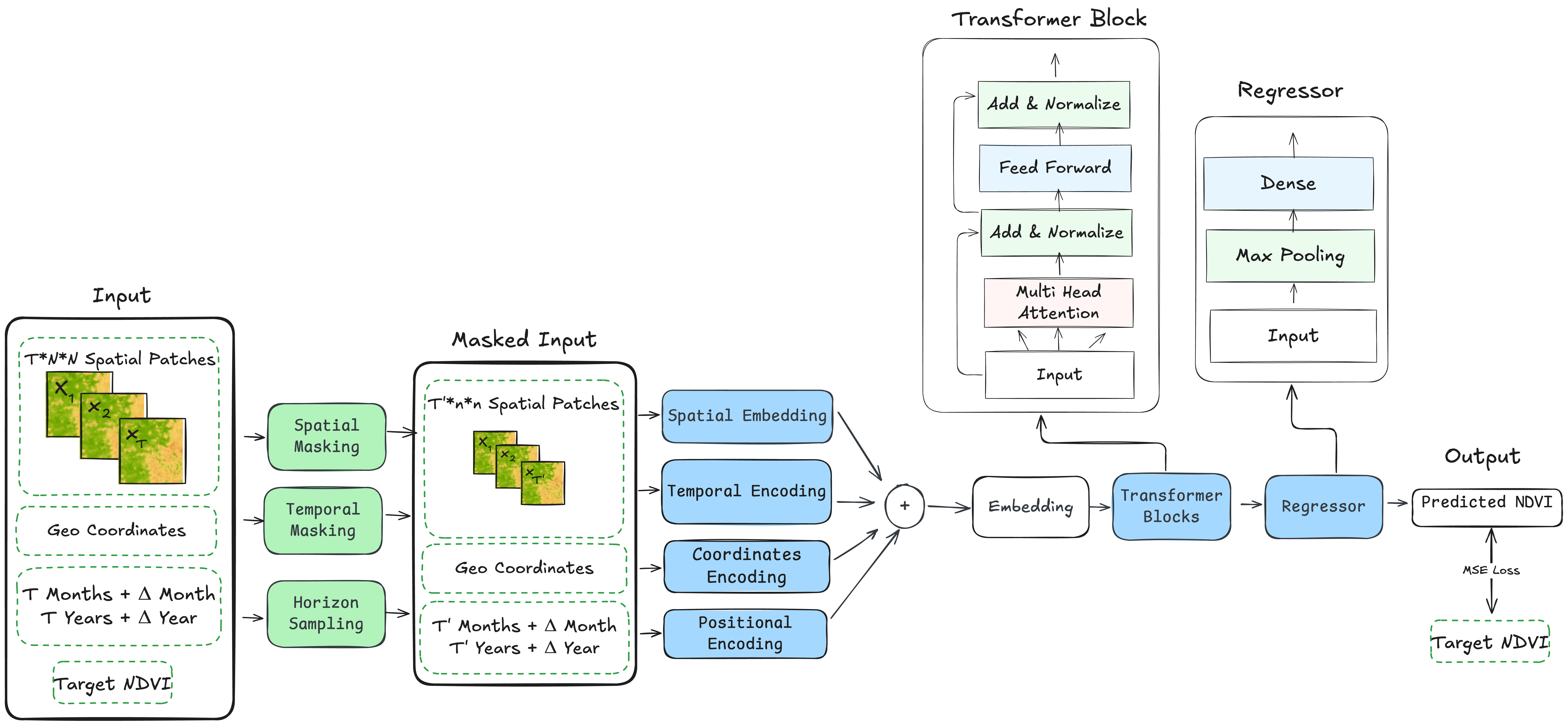}}
\caption{Architecture of the proposed STT-LTF model, combining spatial, temporal, and geographic features. Input spatio-temporal patches undergo masking strategies before multi-channel embedding (spatial, temporal, coordinates, positional), transformer processing, and regression to predict target NDVI values through self-supervised learning.}
\label{fig:transformer_model}
\end{figure*}

Figure \ref{fig:transformer_model} illustrates the architecture of the proposed Transformer-based model for long-term NDVI prediction. The figure shows four main stages:

\begin{itemize}
    \item \textbf{Input Masking}: The input includes spatial NDVI patches, geographic coordinates, and the target future month and year. Spatial and temporal masking, along with horizon sampling, generate diverse training samples.
    
    \item \textbf{Embedding Layers}: The masked inputs are transformed into feature vectors through spatial, temporal, coordinates, and positional embeddings, capturing patterns in both space and time.
    
    \item \textbf{Transformer Encoding}: The combined embeddings are processed through Transformer blocks that model complex spatio-temporal dependencies.
    
    \item \textbf{Regressor Decoder}: The encoded features are pooled and passed through a dense layer to predict the NDVI at the target time, with predictions optimized using MSE loss against ground truth values.
\end{itemize}

\subsection{Input Masking}
The initial stage of our framework involves comprehensive sampling and masking of satellite imagery time series to create diverse training scenarios that enhance model robustness and generalization capabilities. Our approach addresses the inherent challenges in remote sensing data, including irregular temporal sampling, varying spatial scales, and the need for multi-horizon predictions across different forecasting periods.

For each pixel of interest, we construct a structured input sequence as defined in Equation \ref{eqinput}, which encompasses spatial NDVI patches, temporal metadata, and geographic coordinates. This multi-faceted representation enables simultaneous utilization of spatial patterns within local neighborhoods, temporal dynamics across multiple timescales, and geographic context that influences vegetation behavior patterns \cite{gao2022earthformer}.

To enhance model adaptability and prevent overfitting to specific spatiotemporal configurations, we implement a comprehensive masking strategy that introduces structured variability during training. This approach draws inspiration from recent advances in self-supervised learning for time series forecasting and addresses the unique challenges of Earth observation data \cite{zhou2021informer,yao2023ringmo}. The masking strategy consists of three complementary operations applied stochastically during training.

\textbf{Spatial masking} addresses the varying spatial scales encountered in real-world applications by randomly selecting a spatial window size $n$ where $n < N$ and setting all NDVI values outside the central $n \times n$ region to zero. This operation forces the model to adapt to different spatial contexts and scales, which is particularly important given that vegetation patterns exhibit distinct spatial correlation structures at different resolutions \cite{chen2022swinst}. The spatial masking enables the model to learn scale-invariant representations and handle scenarios with limited spatial context availability.

\textbf{Temporal masking} tackles the irregular data availability common in satellite imagery due to cloud cover, sensor malfunctions, or acquisition constraints by randomly selecting a prefix length $T' < T$ and masking all observations and corresponding temporal tokens that occur before time step $T'$. This creates training scenarios where predictions must be made from truncated historical sequences, improving robustness to missing observations and enabling effective learning from sequences of varying lengths \cite{verbesselt2020detecting}. The temporal masking strategy prepares the model for real-world conditions where complete temporal coverage is rarely available.

\textbf{Future horizon sampling} enables direct multi-horizon forecasting by uniformly sampling the prediction horizon $\Delta$ from the range:

\begin{equation}
\Delta \sim \text{Uniform}\{1, 2, \ldots, Y_{\text{max}} - T\}
\end{equation}

where $\Delta$ represents the number of time steps beyond the last observed measurement. This approach exposes the model to both short-term predictions (small $\Delta$) and long-term forecasting scenarios (large $\Delta$) during training, avoiding the error accumulation issues inherent in autoregressive methods \cite{lim2021temporal}. The sampling strategy enables flexible prediction capabilities across temporal scales without requiring iterative inference procedures.

The integrated application of these three masking operations creates a comprehensive data augmentation framework that significantly enhances model robustness across diverse spatiotemporal scales and prediction horizons. By simultaneously varying spatial context, temporal sequence lengths, and forecasting horizons, we establish a training environment that effectively prepares the model for the heterogeneous conditions encountered in operational Earth monitoring applications \cite{tseng2023lightweight}. This masking approach differs fundamentally from traditional computer vision techniques by specifically addressing the unique challenges of spatiotemporal Earth observation data, resulting in a unified model capable of generalizing across diverse spatial scales, temporal contexts, and forecasting requirements.

\subsection{Observation Embedding}

The observation embedding layer transforms the input sequence into a high-dimensional feature space while preserving the relationships between spatial, temporal, and geographic components. The input sequence is defined as:

\begin{equation}
\begin{split}
O &= \{(X_t, M_t, Y_t, \text{lat}, \text{lon}) \mid t = 0, \ldots, T\} \\
&\quad \cup \{(M_{T+\Delta}, Y_{T+\Delta})\}
\end{split}
\end{equation}

The embedding process consists of four main components:

\subsubsection{NDVI Spatial Embedding}

NDVI patches $\{X_t\}_{t=0}^T$ where each $X_t \in \mathbb{R}^{N \times N}$ are flattened and projected into a higher-dimensional vector space via a linear layer. This captures spatial variability around each target pixel:

\begin{equation}
E_{\text{spatial}}(X_t) = \text{Linear}(\text{flatten}(X_t))
\end{equation}

where $\text{Linear}: \mathbb{R}^{N^2} \rightarrow \mathbb{R}^{d_{\text{embed}}}$ maps the flattened spatial patch to the embedding dimension.

\subsubsection{Cyclical Temporal Encoding}

Temporal data $(\{M_t, Y_t\}_{t=0}^T)$ requires special handling to capture both seasonal cycles and long-term trends. Monthly information uses cyclical encoding to ensure temporal continuity:

\begin{equation}
E_{\text{month}}(M_t) = \left[\sin\left(\frac{2\pi M_t}{12}\right), \cos\left(\frac{2\pi M_t}{12}\right)\right]
\end{equation}

Years are normalized relative to the dataset bounds and linearly projected:

\begin{equation}
Y_{\text{norm}} = \frac{Y_t - Y_{\text{start}}}{Y_{\text{end}} - Y_{\text{start}}}
\end{equation}

\begin{equation}
E_{\text{year}}(Y_t) = \text{Linear}(Y_{\text{norm}})
\end{equation}

The target month $(M_{T+\Delta})$ and year $(Y_{T+\Delta})$ are encoded identically and appended to indicate the prediction timestamp.

\subsubsection{Location Encoding}

Latitude and longitude coordinates are converted to 3D Cartesian coordinates on a unit sphere to preserve geographic relationships:

\begin{equation}
\begin{aligned}
x &= \cos(\text{lat}) \cdot \cos(\text{lon}) \\
y &= \cos(\text{lat}) \cdot \sin(\text{lon}) \\
z &= \sin(\text{lat})
\end{aligned}
\end{equation}

The 3D coordinates are then projected into the embedding space:

\begin{equation}
E_{\text{location}}(\text{lat}, \text{lon}) = \text{Linear}([x, y, z])
\end{equation}

\subsubsection{Positional Encoding}

Sinusoidal positional encoding preserves the sequential order of observations:

\begin{equation}
\begin{split}
PE(t) &= [\sin(2\pi k t), \cos(2\pi k t) \\
&\quad \mid k = 0, \ldots, \lfloor d/2 \rfloor]
\end{split}
\end{equation}

where $d$ is the embedding dimension and $k$ indexes different frequency components.

\subsubsection{Final Embedding Concatenation}
For each time step $t$, the individual embeddings are concatenated and projected into the model dimension:

\begin{equation}
\begin{split}
R_t &= \text{concat}(E_{\text{spatial}}(X_t), E_{\text{month}}(M_t), \\
&\quad E_{\text{year}}(Y_t), E_{\text{location}}(\text{lat}, \text{lon}))
\end{split}
\end{equation}

\begin{equation}
H_t = \text{Linear}(R_t)
\end{equation}

Finally, positional encoding is added via element-wise summation:

\begin{equation}
O_t = H_t + PE(t)
\end{equation}

This formulation captures spatial patterns, temporal dynamics, geographic context, and sequential structure necessary for accurate spatio-temporal NDVI prediction.

\subsection{Transformer Encoder}
The embedded sequence $\{O_t\}_{t=0}^{T+1}$ is processed through a stack of $L$ transformer encoder blocks, each consisting of multi-head self-attention and feed-forward components. Following the architecture principles of BERT \cite{devlin2018bert}, each transformer block applies residual connections and layer normalization to ensure stable training and effective information flow.

For each transformer layer $l$, the multi-head attention mechanism computes attention weights across all time steps simultaneously. Given the input representation $H^{(l-1)} \in \mathbb{R}^{(T+2) \times d}$ from the previous layer, the attention computation is:

\begin{equation}
\text{Attention}(Q, K, V) = \text{softmax}\left(\frac{QK^T}{\sqrt{d_k}}\right)V
\end{equation}

where $Q$, $K$, and $V$ are the query, key, and value matrices, respectively, computed as:

\begin{equation}
\begin{aligned}
Q &= H^{(l-1)}W_Q \\
K &= H^{(l-1)}W_K \\
V &= H^{(l-1)}W_V
\end{aligned}
\end{equation}

The multi-head mechanism applies $h$ parallel attention operations and concatenates their outputs:

\begin{equation}
\begin{split}
\text{MultiHead}(H^{(l-1)}) &= \text{concat}(\text{head}_1, \ldots, \text{head}_h)W_O \\
\text{where } \text{head}_i &= \text{Attention}(Q_i, K_i, V_i)
\end{split}
\end{equation}

Each transformer block applies a position-wise feed-forward network followed by residual connections and layer normalization:

\begin{equation}
\begin{aligned}
\text{FFN}(x) &= \text{ReLU}(xW_1 + b_1)W_2 + b_2 \\
H^{(l)} &= \text{LayerNorm}(H^{(l-1)} + \text{MultiHead}(H^{(l-1)})) \\
H^{(l)} &= \text{LayerNorm}(H^{(l)} + \text{FFN}(H^{(l)}))
\end{aligned}
\end{equation}

The stacked architecture enables hierarchical feature learning, where early layers capture local temporal patterns while deeper layers learn global relationships and long-term trends. The final output $H^{(L)}$ provides rich spatio-temporal representations that integrate local variations, long-range dependencies, spatial context, and geographic relationships necessary for accurate NDVI prediction.

\subsection{Regression Decoder and Self-Supervised Training}

The final transformer encoding corresponding to the target timestamp is processed through a regression decoder to produce the NDVI prediction for the center pixel. This decoder consists of a two-layer fully connected network with ReLU activation and dropout regularization:

\begin{equation}
\begin{split}
h &= \text{ReLU}(\text{Linear}_1(H^{(L)}_{T+\Delta})) \\
\hat{x}_{T+\Delta} &= \text{Linear}_2(\text{Dropout}(h))
\end{split}
\end{equation}

where $\text{Linear}_1: \mathbb{R}^{d_{\text{model}}} \rightarrow \mathbb{R}^{d_{\text{model}}/2}$ reduces the dimensionality, $H^{(L)}_{T+\Delta}$ represents the transformer encoding at the target time step, and $\text{Linear}_2: \mathbb{R}^{d_{\text{model}}/2} \rightarrow \mathbb{R}$ maps to a single scalar NDVI value for the center pixel.

The model employs self-supervised learning by leveraging the temporal structure of satellite imagery data. The training objective minimizes the discrepancy between predicted and observed NDVI values using Mean Absolute Error (MAE) loss:

\begin{equation}\label{mae}
L(\hat{x}_{T+\Delta}, x_{T+\Delta}) = |\hat{x}_{T+\Delta} - x_{T+\Delta}|
\end{equation}

Parameter optimization is performed through backpropagation using the Adam optimizer. This self-supervised approach enables the model to learn complex spatio-temporal patterns without requiring manually labeled data, effectively capturing dependencies across spatial neighborhoods, seasonal cycles, and long-term environmental trends.

\section{\uppercase{Experimental Results}}
\subsection{\uppercase{Study Area}}

Our research focuses on the southeastern Mediterranean basin, located along a gradient transitioning from Mediterranean to arid climatic conditions (Figure~\ref{fig:study_area_combined}). The rainfall varies between 450 mm/year to 250 mm/year, resulting in a transition from shrublands to phrygana (Bata) to bare desert.

The study region presents unique challenges for vegetation monitoring due to its inherent environmental heterogeneity. Precipitation patterns exhibit considerable inter-annual variability, with recurring drought cycles and periodic disturbances from wildfires significantly influencing landscape dynamics. Human activities further contribute to ecosystem modifications, resulting in a complex spatial pattern of vegetation, exposed soil, and rocky surfaces. These combined factors create substantial temporal variability in vegetation responses, as demonstrated by the seasonal and inter-annual NDVI fluctuations observed throughout the study period.

We evaluated the proposed method using multi-mission Landsat imagery spanning four decades (1984-2024), incorporating data from Landsat Missions 5, 7, 8, and 9. This comprehensive dataset represents Mediterranean ecosystems characterized by pronounced spatial heterogeneity and temporal dynamics, including extended dry periods punctuated by intense precipitation events as noted in \cite{faran2020}.

The spatial extent covers $53.94 \times 37.35$ $\rm{km}^2$, represented as a $1798 \times 1245$ pixel array with 30-meter spatial resolution. Source imagery was acquired through Google Earth Engine's Level-2 surface reflectance products. To address cloud contamination and seasonal variations, we generated bi-annual composite images by calculating median NDVI values from cloud-free observations ($<$ 20\%  cloud cover) for two distinct periods: October through April, representing the "Winter" growing season, and May through September, representing the "Summer" dormant season.

The complete dataset encompasses 2,238,510 individual pixel time series. For model development and evaluation, we implemented a temporal split strategy: observations from 1984-2013 (spanning 30 years) were allocated for model training and validation, while the period from 2014-2024 served as an independent test set for assessing temporal generalization capabilities. This approach ensures robust evaluation of long-term prediction performance on genuinely unseen temporal data.

\begin{figure*}[htb] 
\centering 
\includegraphics[width=1\textwidth, height=0.4\textheight]{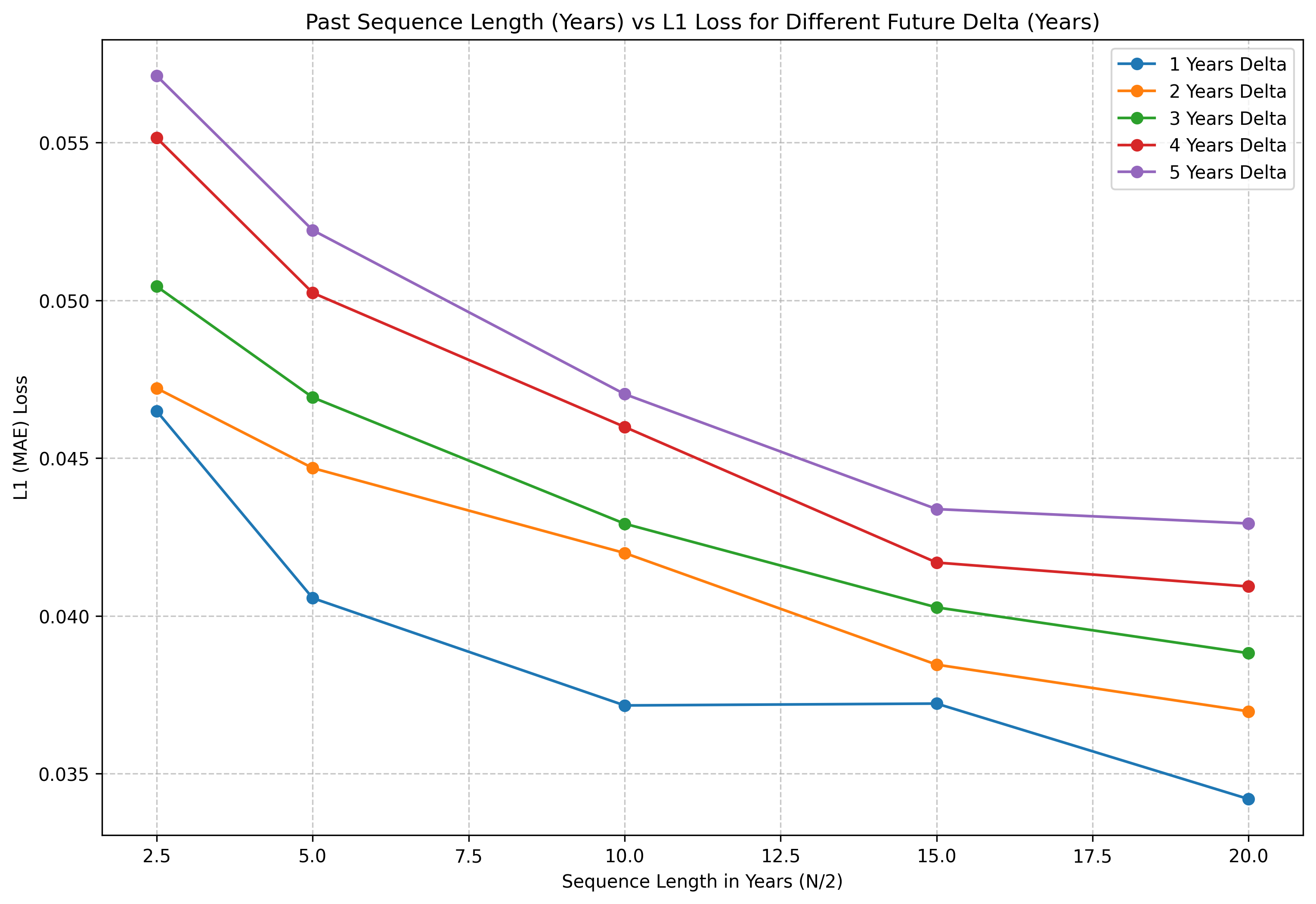} 
\caption{MAE loss against past sequence length for different future prediction horizons ($\Delta t$) with spatial patch size 1. Results show that MAE loss decreases with longer historical sequences and shorter prediction horizons. Adopted from \cite{faran2025sstltp}.} 
\label{fig:results} 
\end{figure*}

\begin{figure*}[t] \centering \includegraphics[width=1\textwidth, height=0.9\textheight]{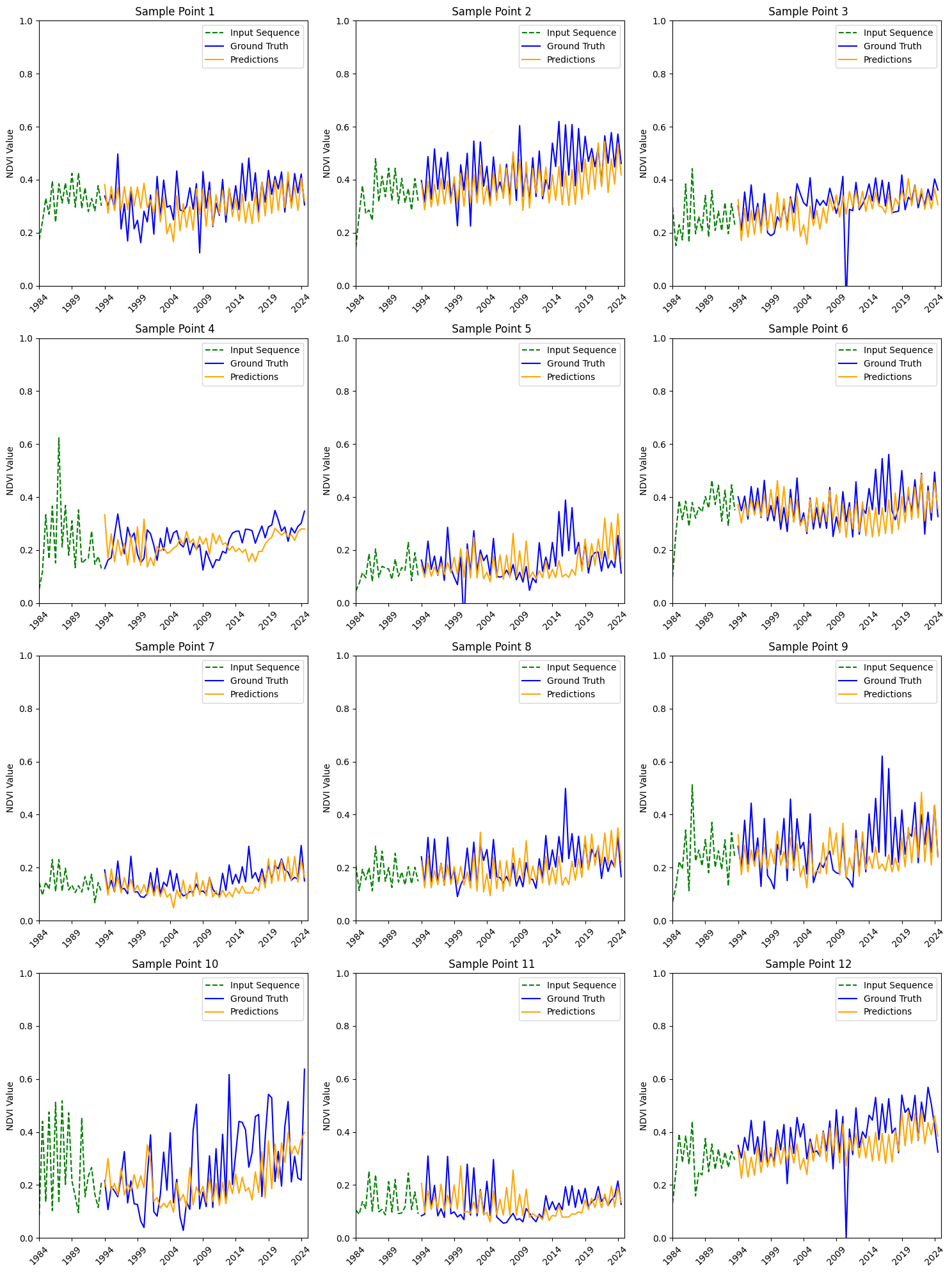} \caption{Performance comparison across 12 sample points using 10-year input sequences. Results demonstrate accurate prediction of seasonal patterns and long-term trends. Adopted from \cite{faran2025sstltp}.}
 \label{fig:prediction2} \end{figure*}

\subsection{Implementation and Parameters}
Our model employs a multi-dimensional embedding architecture with three distinct embedding layers: a 128-dimensional layer for NDVI spatial features, an 8-dimensional layer for temporal information (year and season), and an 8-dimensional layer for geographic coordinates (latitude and longitude). These embeddings are combined and projected through a linear layer to a unified 256-dimensional space, which serves as input to the transformer architecture~\cite{vaswani2017attention,tseng2023lightweight}.

The transformer backbone consists of three encoding blocks, each equipped with eight attention heads to capture diverse spatial and temporal dependencies. A dropout rate of 0.2 is applied after the embedding layer and each transformer block to prevent overfitting~\cite{gao2022earthformer}. The model was trained for 150 epochs using the Adam optimizer with an initial learning rate of $2 \times 10^{-4}$.

For the loss function, we employ Mean Absolute Error (MAE), as defined in Equation~\ref{mae}, which was selected due to its robustness to outliers in satellite time series data~\cite{gomez2016optical}. Model performance is evaluated using both MAE and the coefficient of determination ($R^2$) to provide comprehensive assessment of prediction accuracy and variance explanation capability.

\begin{table*}[ht]
\centering
\begin{tabular}{| l || c | c | c | c | c |}
\hline
Spatial/Sequence & Sequence 1 & Sequence 5 & Sequence 10 & Sequence 15 & Sequence 20 \\ \hline
Patch 1×1 & MAE: 0.096 & MAE: 0.098 & MAE: 0.070 & MAE: 0.057 & MAE: 0.025 \\
          & R\textsuperscript{2}: 0.092 & R\textsuperscript{2}: 0.225 & R\textsuperscript{2}: 0.535 & R\textsuperscript{2}: 0.678 & R\textsuperscript{2}: 0.922 \\ \hline
Patch 3×3 & MAE: 0.098 & MAE: 0.101 & MAE: 0.072 & MAE: 0.054 & MAE: 0.024 \\
          & R\textsuperscript{2}: 0.042 & R\textsuperscript{2}: 0.179 & R\textsuperscript{2}: 0.525 & R\textsuperscript{2}: 0.706 & R\textsuperscript{2}: 0.926 \\ \hline
Patch 5×5 & MAE: 0.098 & MAE: 0.102 & MAE: 0.073 & MAE: 0.051 & MAE: 0.025 \\
          & R\textsuperscript{2}: 0.048 & R\textsuperscript{2}: 0.166 & R\textsuperscript{2}: 0.507 & R\textsuperscript{2}: 0.723 & R\textsuperscript{2}: 0.924 \\ \hline
Patch 7×7 & MAE: 0.098 & MAE: 0.104 & MAE: 0.073 & MAE: 0.050 & \textbf{MAE: 0.024} \\
          & R\textsuperscript{2}: 0.044 & R\textsuperscript{2}: 0.147 & R\textsuperscript{2}: 0.515 & R\textsuperscript{2}: 0.736 & \textbf{R\textsuperscript{2}: 0.927} \\ \hline
Patch 9×9 & MAE: 0.098 & MAE: 0.104 & MAE: 0.071 & MAE: 0.049 & MAE: 0.024 \\
          & R\textsuperscript{2}: 0.027 & R\textsuperscript{2}: 0.143 & R\textsuperscript{2}: 0.531 & R\textsuperscript{2}: 0.747 & R\textsuperscript{2}: 0.925 \\ \hline
\end{tabular}
\caption{\label{tab:metrics}Mean Absolute Error (MAE) and R² performance across different spatial patch sizes and temporal sequence lengths}
\end{table*}

\subsection{Temporal Evaluation}
We evaluated the proposed model's prediction capabilities using various sequence lengths ($T$) and future prediction horizons ($\Delta$) with the spatial patch size ($N$) set to 1. This assessment helped determine our method's ability to capture time-series patterns and seasonality across different temporal scales. The results are presented in Figure~\ref{fig:results}.

A past observation sequence of 10 years yielded the most significant improvement in prediction accuracy, showing noticeable gains compared to shorter sequences. Beyond 10 years, further improvements are less pronounced, with 15 to 20 years achieving the lowest $L_1$ loss. The model performs optimally for next-year predictions, with accuracy gradually declining as the prediction horizon extends.

Analyzing results by land-cover class might reveal different patterns, as some classes (e.g., building areas, bare ground) exhibit static NDVI values, while others (e.g., grass, shrub) display greater temporal variability.

Figure~\ref{fig:prediction2} demonstrates the model's output using a 10-year input sequence applied to sample locations. The model successfully captures overall NDVI trends over time, aligning well with observed patterns in the input sequence. However, it struggles to predict unexpected outliers that may result from sudden environmental events or changes. These deviations highlight the inherent challenges of modeling abrupt anomalies within a predominantly trend-based framework.

\subsection{Spatial Evaluation}
We also conducted a comprehensive analysis examining how spatial coverage, represented by varying patch dimensions, influences model prediction performance. The experimental setup employed sequence lengths of 10 years with single-year prediction horizons, as presented in Table~\ref{tab:metrics}. Our findings demonstrate that expanding the spatial window beyond $7 \times 7$ pixels (approximately $210\,\text{m} \times 210\,\text{m}$) produced marginal improvements in prediction accuracy.

This spatial threshold suggests that the optimal window size effectively captures the spectral characteristics of uniform vegetation communities within our study area~\cite{zhang2024reconstruction}. Beyond this scale, the inclusion of additional pixels primarily incorporates adjacent areas with distinct land cover characteristics, potentially introducing spectral heterogeneity that dilutes the vegetation signal~\cite{rosberg2022estimating,vanleeuwen2006comparison}. This phenomenon aligns with established remote sensing principles, where mixed pixels containing multiple cover types can compromise classification accuracy and reduce the strength of vegetation indices~\cite{singh2024monitoring}.

The computational efficiency also supports this finding, as larger spatial windows require significantly more processing resources while providing diminishing returns in predictive accuracy~\cite{verma2023vegetation}. This trade-off between computational cost and model performance suggests that the $7 \times 7$ pixel configuration represents an optimal balance for operational vegetation monitoring applications~\cite{cheng2024combination}.

In contrast to the limited spatial benefits observed beyond the optimal window size, temporal sequence length demonstrated substantially greater influence on model performance. Extended temporal sequences consistently improved both Mean Absolute Error (MAE) and coefficient of determination (R\textsuperscript{2}) metrics across all spatial configurations. This temporal advantage suggests that historical context provides more valuable predictive information than expanding spatial coverage, enabling the model to capture long-term vegetation trends, seasonal cycles, and inter-annual variability patterns more effectively~\cite{chu2021long}.

Our results emphasize the importance of balancing spatial and temporal dimensions in vegetation prediction models. While higher spatial resolution imagery could potentially reveal finer-scale patterns, the current analysis suggests that temporal depth offers greater predictive value than spatial extent at our operational scale~\cite{sharifi2021enhanced}.

\subsection{Comparison with Other Models}
We evaluated our proposed model against several established time-series analysis approaches, including Support Vector Machines (SVM), Fully-Connected Neural Networks, 1D Convolutional Neural Networks (CNN-1D), Long Short-Term Memory (LSTM) networks, and the Temporal Fusion Transformer (TFT) \cite{lim2021temporal}. 

The SVM implementation utilized a Radial Basis Function (RBF) kernel, a standard choice for time-series regression tasks. The Fully-Connected Neural Network comprised three layers with 128 hidden neurons each, using ReLU activation functions to process the flattened 10-sample input sequences. The CNN-1D architecture employed three convolutional layers with 64 filters and kernel size 3, maintaining sequence length through padding before the final linear layer. The LSTM model featured a 3-layer stacked design with 64 hidden units per layer to capture sequential temporal patterns.

For the TFT implementation, we adapted the architecture to NDVI prediction by configuring it with an embedding dimension of 128, four attention heads, and three transformer blocks. The model incorporated temporal encodings for seasonal patterns.

The results presented in Table \ref{tab:tablePerformance} reveal the comparative performance of our STT-LTF model. A notable advantage of our approach lies in its architectural design, which accepts target year information as direct input, enabling multi-horizon predictions without iterative inference. Traditional autoregressive methods require sequential predictions for long-term forecasting, where each predicted value serves as input for subsequent predictions. This iterative process leads to error accumulation and increased computational overhead with each step. In contrast, our direct prediction capability overcomes these limitations inherent in autoregressive approaches, while achieving superior accuracy for long-term NDVI forecasting tasks.

\begin{table}[bh]
\begin{center}
\begin{tabular}{| c || c | c |}
\hline
\textbf{Model} & \textbf{MAE} & \textbf{R\textsuperscript{2}} \\ [0.5ex]
\hline\hline
SVM & 0.0456 & 0.7459 \\
\hline
CNN-1D & 0.0391 & 0.7810 \\
\hline
Fully Connected & 0.0383 & 0.7953 \\
\hline
LSTM & 0.0363 & 0.8088 \\
\hline
Transformer & 0.0361 & 0.8182 \\
\hline
\textbf{Proposed Model} & \textbf{0.0328} & \textbf{0.8412} \\
\hline \hline
\end{tabular}
\end{center}
\caption{\label{tab:tablePerformance} Performance comparison of previous methods and our STT-LTF model for next-year prediction using 10-year input sequences..}
\end{table}

\section{\uppercase{Conclusion}}
This research presented STT-LTF, a self-supervised spatio-temporal transformer framework designed for multi-horizon vegetation forecasting using satellite observations. The proposed methodology addresses fundamental challenges in environmental monitoring by simultaneously processing spatial neighborhoods and temporal sequences without requiring labeled training data. Through comprehensive masking strategies and direct prediction mechanisms, our framework learns meaningful representations from four decades of Landsat imagery, capturing both local vegetation patterns and regional climate influences across Mediterranean ecosystems.
The experimental evaluation revealed important insights into the tradeoff between spatial and temporal dimensions in vegetation modeling. While expanding spatial context showed diminishing returns beyond a certain threshold, temporal depth consistently enhanced prediction accuracy across various configurations. The direct prediction approach eliminated error accumulation inherent in autoregressive methods, enabling accurate forecasts at arbitrary future timestamps through a single forward pass.

STT-LTF demonstrated superior performance compared to various established approaches spanning traditional machine learning methods and modern deep learning architectures. The framework's ability to process variable-length sequences with missing observations makes it particularly suitable for operational Earth monitoring applications where data availability varies. By learning from unlabeled historical archives, the approach overcomes the scarcity of annotated multi-decade datasets while maintaining prediction accuracy across heterogeneous landscapes.

Future research directions include extending the framework to incorporate multi-modal satellite data, integrating climate variables and human activity indicators, and developing methods for geographic transferability to different ecosystems and climatic zones. Additionally, implementing more efficient computational approaches, such as Vision Transformer architectures, would enable processing larger geographic areas and longer temporal sequences without excessive memory requirements, making the framework practical for large-scale environmental monitoring systems. Advanced mechanisms that identify and prioritize threshold areas with a higher likelihood of vegetation change could enhance prediction accuracy in critical transition zones, particularly in regions vulnerable to desertification or rapid land use transformation.

\clearpage

\bibliographystyle{apalike}
{\small
\bibliography{bibliography}}

@article{kennedy2018,
title={Bringing an ecological view of change to {L}andsat‐based remote sensing},
author={Kennedy, R. E. and Andréfouët, S. and Cohen, W. B. and Gómez, C. and Griffiths, P. and Hais, M. and others},
journal={Frontiers in Ecology and the Environment},
volume={16},
number={6},
pages={340--348},
year={2018}
}

@article{zhu2017change,
title={Change detection using {Landsat} time series: A review of frequencies, preprocessing, algorithms, and applications},
author={Zhu, Z.},
journal={Journal of Photogrammetry and Remote Sensing},
volume={130},
pages={370--384},
year={2017}
}

@article{zhu2019,
title={Continuous monitoring of land disturbance based on {L}andsat time series},
author={Zhu, Z. and Zhang, J. and Yang, Z. and Aljaddani, A. H. and Cohen, W. B. and Qiu, S. and Zhou, C.},
journal={Remote Sensing of Environment},
volume={238},
pages={111116},
year={2019}
}

@article{miller2024,
  title={Deep learning for satellite image time-series analysis: A review},
  author={Miller, Lynn and Pelletier, Charlotte and Webb, Geoffrey I},
  journal={IEEE Geoscience and Remote Sensing Magazine},
  year={2024},
  publisher={IEEE}
}

@article{yuan2020,
title={Self-Supervised Pretraining of Transformers for Satellite Image Time Series Classification},
author={Yuan, Y. and Lin, L.},
journal={IEEE Journal of Selected Topics in Applied Earth Observations and Remote Sensing},
volume={14},
pages={474--487},
year={2020},
publisher={IEEE}
}

@article{russwurm2018,
title={Self-attention for raw optical satellite time series classification},
author={Ru{\ss}wurm, M. and Körner, M.},
journal={Journal of Photogrammetry and Remote Sensing},
volume={169},
pages={421--435},
year={2018},
publisher={Elsevier}
}

@inproceedings{faran2020,
  title={Multi Seasonal Deep Learning Classification of {VEN}u{S} Images},
  author={Faran, Ido and Netanyahu, Nathan S and David, Eli and Rud, Ronit and Shoshany, Maxim},
  booktitle={Proceedings of the IEEE International Geoscience and Remote Sensing Symposium},
  pages={6754--6757},
  year={2020}
}

@inproceedings{devlin2018bert,
  title={Bert: Pre-training of deep bidirectional transformers for language understanding},
  author={Devlin, Jacob and Chang, Ming-Wei and Lee, Kenton and Toutanova, Kristina},
  booktitle={Proceedings of the 2019 conference of the North American chapter of the association for computational linguistics: human language technologies, volume 1 (long and short papers)},
  pages={4171--4186},
  year={2019}
}

@inproceedings{vaswani2017attention,
  title={Attention is all you need},
  author={Vaswani, A and Shazeer, N. and Parmar, N. and Uszkoreit, J. and Jones, L. and Gomez, A. N. and Keiser, L. and Polosukhin, I.},
  booktitle={Advances in Neural Information Processing Systems},
  pages={6000--6010},
  year={2017}
}

@article{taylor2018forecasting,
  title={Forecasting at scale},
  author={Taylor, Sean J and Letham, Benjamin},
  journal={The American Statistician},
  volume={72},
  number={1},
  pages={37--45},
  year={2018},
  publisher={Taylor \& Francis}
}

@book{box2015time,
  title={Time Series Analysis: Forecasting and Control},
  author={Box, George EP and Jenkins, Gwilym M and Reinsel, Gregory C and Ljung, Greta M},
  year={2015},
  publisher={John Wiley \& Sons}
}

@inproceedings{zhou2021informer,
  title={Informer: Beyond efficient transformer for long sequence time-series forecasting},
  author={Zhou, Haoyi and Zhang, Shanghang and Peng, Jieqi and Zhang, Shuai and Li, Jianxin and Xiong, Hui and Zhang, Wancai},
  booktitle={Proceedings of the AAAI Conference on Artificial Intelligence},
  volume={35},
  pages={11106--11115},
  year={2021}
}

@article{ruiz2014hybrid,
  title={Hybrid approaches based on SARIMA and artificial neural networks for inspection time series forecasting},
  author={Ruiz-Aguilar, JJ and Turias, IJ and Jim{\'e}nez-Come, MJ},
  journal={Transportation Research Part E: Logistics and Transportation Review},
  volume={67},
  pages={1--13},
  year={2014},
  publisher={Elsevier}
}

@article{lim2021temporal,
  title={Temporal fusion transformers for interpretable multi-horizon time series forecasting},
  author={Lim, Bryan and Ar{\i}k, Sercan {\"O} and Loeff, Nicolas and Pfister, Tomas},
  journal={International Journal of Forecasting},
  volume={37},
  number={4},
  pages={1748--1764},
  year={2021},
  publisher={Elsevier}
}

@inproceedings{roitberg2024primary,
  title={Primary Productivity and Woody growth: a 35 years {L}andsat {TM} {NDVI} time series investigation across desert-fringe in the south-eastern {M}editerranean},
  author={Roitberg, Elena and Shoshany, Maxim},
  booktitle={Proceedings of the IEEE International Geoscience and Remote Sensing Symposium},
  pages={2867--2870},
  year={2024}
}

@inproceedings{zhang2023crossformer,
  title={Crossformer: Transformer utilizing cross-dimension dependency for multivariate time series forecasting},
  author={Zhang, Yunhao and Yan, Junchi},
  booktitle={Proceedings of the Eleventh International Conference on Learning Representations},
  year={2023}
}

@article{liu2023itransformer,
  title={i{T}ransformer: Inverted {T}ransformers are effective for time series forecasting},
  author={Liu, Yong and Hu, Tengge and Zhang, Haoran and Wu, Haixu and Wang, Shiyu and Ma, Lintao and Long, Mingsheng},
  journal={arXiv:2310.06625},
  year={2023}
}

@article{yao2023ringmo,
  title={Ring{M}o-{S}ense: Remote sensing foundation model for spatiotemporal prediction via spatiotemporal evolution disentangling},
  author={Yao, Fanglong and Lu, Wanxuan and Yang, Heming and Xu, Liangyu and Liu, Chenglong and Hu, Leiyi and Yu, Hongfeng and Liu, Nayu and Deng, Chubo and Tang, Deke and Chen, C and Yu, J and Sun, X and Fu, K},
  journal={IEEE Transactions on Geoscience and Remote Sensing},
  volume={61},
  pages = {1--21},
  year={2023},
  publisher={IEEE}
}

@inproceedings{gao2022earthformer,
  title={Earthformer: Exploring space-time transformers for {Earth} system forecasting},
  author={Gao, Zhihan and Shi, Xingjian and Wang, Hao and Zhu, Yi and Wang, Yuyang Bernie and Li, Mu and Yeung, Dit-Yan},
  booktitle={Advances in Neural Information Processing Systems},
  volume={35},
  pages={25390--25403},
  year={2022}
}

@article{verbesselt2010detecting,
  title={Detecting trend and seasonal changes in satellite image time series},
  author={Verbesselt, Jan and Hyndman, Rob and Newnham, Glenn and Culvenor, Darius},
  journal={Remote sensing of Environment},
  volume={114},
  number={1},
  pages={106--115},
  year={2010},
  publisher={Elsevier}
}

@article{gomez2016optical,
title = {Optical remotely sensed time series data for land cover classification: A review},
journal = {ISPRS Journal of Photogrammetry and Remote Sensing},
volume = {116},
pages = {55-72},
year = {2016},
issn = {0924-2716},
doi = {https://doi.org/10.1016/j.isprsjprs.2016.03.008},
url = {https://www.sciencedirect.com/science/article/pii/S0924271616000769},
author = {Cristina Gómez and Joanne C. White and Michael A. Wulder},
}

@conference{faran2025sstltp,
author={Ido Faran and Nathan S. Netanyahu and Elena Roitberg and Maxim Shoshany},
title={Self-Supervised Transformers for Long-Term Prediction of {Landsat} {NDVI} Time Series},
booktitle={Proceedings of the 14th International Conference on Pattern Recognition Applications and Methods - ICPRAM},
year={2025},
pages={542-552},
publisher={SciTePress},
organization={INSTICC},
doi={10.5220/0013381700003905},
isbn={978-989-758-730-6},
issn={2184-4313},
}

@article{xu2024anthropogenic,
  title={Anthropogenic activities dominated the spatial and temporal changes of normalized difference vegetation index (NDVI) in the Hehuang valley in the northeastern Qinghai Province between 2000 and 2020},
  author={Xu, Bin and Mao, Xufeng and Li, Xingyue and Wei, Xiaoyan and Zhang, Ziping and Tang, Wenjia and Yu, Hongyan and Wu, Yi},
  journal={Frontiers in Environmental Science},
  volume={12},
  pages={1384032},
  year={2024},
  publisher={Frontiers Media SA}
}

@article{chen2024anthropogenic,
  title={Anthropogenic activities dominated the spatial and temporal changes of normalized difference vegetation index ({NDVI}) in the {Hehuang} valley in the northeastern {Qinghai Province} between 2000 and 2020},
  author={Chen, X and others},
  journal={Frontiers in Environmental Science},
  volume={12},
  pages={1384032},
  year={2024},
  publisher={Frontiers}
}

@article{li2021long,
  title={Long time-series {NDVI} reconstruction in cloud-prone regions via spatio-temporal tensor completion},
  author={Li, Y and others},
  journal={Remote Sensing of Environment},
  volume={264},
  pages={112632},
  year={2021},
  publisher={Elsevier}
}

@article{mehmood2024analyzing,
  title={Analyzing vegetation health dynamics across seasons and regions through {NDVI} and climatic variables},
  author={Mehmood, Kaleem and Anees, Shoaib Ahmad and Muhammad, Sultan and Hussain, Khadim and Shahzad, Fahad and Liu, Qijing and Ansari, Mohammad Javed and Alharbi, Sulaiman Ali and Khan, Waseem Razzaq},
  journal={Scientific Reports},
  volume={14},
  number={1},
  pages={11775},
  year={2024},
  publisher={Nature Publishing Group}
}

@article{zhang2024reconstruction,
  title={Reconstruction of dense time series high spatial resolution {NDVI} data using a spatiotemporal optimal weighted combination estimation model based on {Sentinel-2} and {MODIS}},
  author={Zhang, K and Zhu, CM and Li, JL and Shi, KT and Zhang, X},
  journal={Ecological Informatics},
  volume={82},
  pages={102725},
  year={2024},
  publisher={Elsevier}
}

@article{tian2016drought,
  title={Drought forecasting with vegetation temperature condition index using ARIMA models in the Guanzhong Plain},
  author={Tian, Miao and Wang, Peng and Khan, Javed},
  journal={Remote Sensing},
  volume={8},
  number={9},
  pages={690},
  year={2016},
  publisher={MDPI},
  url={https://www.mdpi.com/2072-4292/8/9/690}
}

@article{spatial2024forecasting,
  title={Spatial forecasting of solar radiation using ARIMA model},
  author={Sharma, Nikhil and Sharma, Prashant and Irwin, David and Shenoy, Prashant},
  journal={Remote Sensing Applications: Society and Environment},
  volume={20},
  pages={100423},
  year={2020},
  publisher={Elsevier},
  url={https://www.sciencedirect.com/science/article/abs/pii/S2352938520302731}
}

@article{forecasting2024energy,
  title={Analysis of the Effectiveness of {ARIMA}, {SARIMA}, and {SVR} Models in Time Series Forecasting: A Case Study of Wind Farm Energy Production},
  author={Dudek, Grzegorz and Pe{\l}ka, Pawe{\l} and Smyl, Slawek},
  journal={Energies},
  volume={17},
  number={19},
  pages={4803},
  year={2024},
  publisher={MDPI},
  url={https://www.mdpi.com/1996-1073/17/19/4803}
}

@article{analysis2024effectiveness,
  title={Comparative analysis of SARIMA and deep learning models for short-term offshore wind speed forecasting},
  author={MacKenzie, Cameron and Cannon, Trevor and Garmire, David},
  journal={Energy Reports},
  volume={10},
  pages={2987--2997},
  year={2023},
  publisher={Elsevier},
  url={https://www.sciencedirect.com/science/article/pii/S2352484723011751}
}

@article{zhu2017,
  title={Deep Learning in Remote Sensing: A Comprehensive Review and List of Resources},
  author={Zhu, Xiao Xiang and Tuia, Devis and Mou, Lichao and Xia, Gui-Song and Zhang, Liangpei and Xu, Feng and Fraundorfer, Friedrich},
  journal={IEEE Geoscience and Remote Sensing Magazine},
  volume={5},
  number={4},
  pages={8--36},
  year={2017},
  publisher={IEEE}
}

@article{remotesensingreview2024,
  title={Remote sensing time series analysis: A review of data and applications},
  author={Fu, Yingchun and Zhu, Zhe and Liu, Liangyun and Zhan, Wenfeng and He, Tao and Shen, Huanfeng and Zhao, Jun and Liu, Yongxue and Zhang, Hongsheng and Liu, Zihan and others},
  journal={Journal of Remote Sensing},
  volume={4},
  pages={0285},
  year={2024},
  publisher={AAAS}
}

@article{applicationdeeplearning2023,
  title={Application of deep learning in multitemporal remote sensing image classification},
  author={Cheng, Xinglu and Sun, Yonghua and Zhang, Wangkuan and Wang, Yihan and Cao, Xuyue and Wang, Yanzhao},
  journal={Remote Sensing},
  volume={15},
  number={15},
  pages={3859},
  year={2023},
  publisher={MDPI}
}

@article{transformersreview2024,
  title={Transformers for Remote Sensing: A Systematic Review and Analysis},
  author={He, Guangjun and Johnson, Brian Alan and others},
  journal={Sensors},
  volume={24},
  number={11},
  pages={3495},
  year={2024},
  doi={10.3390/s24113495},
  url={https://pmc.ncbi.nlm.nih.gov/articles/PMC11175147/}
}

@article{changemamba2024,
  title={Changemamba: Remote sensing change detection with spatio-temporal state space model},
  author={Chen, Hongruixuan and Song, Jian and Han, Chengxi and Xia, Junshi and Yokoya, Naoto},
  journal={IEEE Transactions on Geoscience and Remote Sensing},
  year={2024},
  publisher={IEEE}
}

@misc{githubremotesensing2024,
  title={Transformer-based Models for Remote Sensing},
  author={Yang Zhang},
  year={2024},
  howpublished={GitHub repository},
  url={https://github.com/Yangzhangcst/Transformer-in-Computer-Vision/blob/main/main/remote-sensing.md}
}

@article{transformermodels2024,
  title={Transformer models for Land Cover Classification with Satellite Image Time Series},
  author={Voelsen, Mirjana and Rottensteiner, Franz and Heipke, Christian},
  journal={Journal of Photogrammetry, Remote Sensing and Geoinformation Science},
  volume={92},
  number={5},
  pages={547--568},
  year={2024},
  publisher={Springer}
}

@article{foundationmodelssurvey2024,
  title={Foundation Models for Remote Sensing and Earth Observation: A Survey},
  author={Xiao, Aoran and others},
  journal={arXiv preprint},
  year={2024},
  doi={10.48550/arXiv.2410.16602},
  url={https://arxiv.org/abs/2410.16602}
}

@article{visionmamba2025,
  title={Vision Mamba in Remote Sensing: A Comprehensive Survey of Techniques, Applications and Outlook},
  author={Bao, Muyi and others},
  journal={arXiv preprint},
  year={2025},
  doi={10.48550/arXiv.2505.00630},
  url={https://arxiv.org/abs/2505.00630}
}

@article{wang2022selfsupervised,
  title={Self-supervised learning in remote sensing: A review},
  author={Wang, Yi and Albrecht, Conrad M and Braham, Nassim Ait Ali and Mou, Lichao and Zhu, Xiao Xiang},
  journal={IEEE Geoscience and Remote Sensing Magazine},
  volume={10},
  number={4},
  pages={213--247},
  year={2022},
  publisher={IEEE}
}

@article{selfsupreview2022,
  title={Self-supervised learning for scene classification in remote sensing: Current state of the art and perspectives},
  author={Berg, Paul and Pham, Minh-Tan and Courty, Nicolas},
  journal={Remote Sensing},
  volume={14},
  number={16},
  pages={3995},
  year={2022},
  publisher={MDPI}
}

@article{indomain2024,
  title={Self-supervised in-domain representation learning for remote sensing image scene classification},
  author={Ghanbarzadeh, Ali and Soleimani, Hossein},
  journal={Heliyon},
  volume={10},
  number={19},
  year={2024},
  publisher={Elsevier}
}

@article{tseng2023lightweight,
  title={Lightweight, Pre-trained Transformers for Remote Sensing Timeseries},
  author={Tseng, Gabriel and others},
  journal={arXiv preprint},
  year={2023}
}

@article{chen2022swinst,
  title={SwinSTFM: Remote sensing spatiotemporal fusion using swin transformer},
  author={Chen, Guanyu and Jiao, Qingsong and Hu, Qihao and Xiao, Liang and Ye, Zheng},
  journal={IEEE Transactions on Geoscience and Remote Sensing},
  volume={60},
  pages={1--18},
  year={2022},
  url={https://ieeexplore.ieee.org/document/9761206}
}

@article{verbesselt2020detecting,
  title={Detecting trend and seasonal changes in satellite image time series},
  author={Verbesselt, Jan and Hyndman, Rob and Newnham, Glenn and Culvenor, Darius},
  journal={Remote Sensing of Environment},
  volume={114},
  number={1},
  pages={106--115},
  year={2010},
  url={https://www.sciencedirect.com/science/article/pii/S003442570900265X}
}

@article{rosberg2022estimating,
  title={Estimating vegetation indices and biophysical parameters for Central European temperate forests with Sentinel-1 SAR data and machine learning},
  author={Ro{\ss}berg, N and Schmitt, M},
  journal={Taylor \& Francis Online},
  year={2024},
  url={https://www.tandfonline.com/doi/full/10.1080/20964471.2025.2459300}
}

@article{vanleeuwen2006comparison,
  title={A commentary review on the use of normalized difference vegetation index (NDVI) in the era of popular remote sensing},
  author={Van Leeuwen, WJD and Huete, AR and Laing, TW},
  journal={Journal of Forestry Research},
  year={2020},
  url={https://link.springer.com/article/10.1007/s11676-020-01155-1}
}

@article{singh2024monitoring,
  title={Monitoring vegetation degradation using remote sensing and machine learning over India - a multi-sensor, multi-temporal and multi-scale approach},
  author={Singh, M and Arshad, A and Bijlwan, A and Tamang, M and Shahina, NN and Biswas, A},
  journal={Frontiers in Forests and Global Change},
  volume={7},
  year={2024},
  url={https://www.frontiersin.org/journals/forests-and-global-change/articles/10.3389/ffgc.2024.1382557/full}
}

@article{verma2023vegetation,
  title={Advances in vegetation mapping through remote sensing and machine learning techniques: a scientometric review},
  author={Verma, A and Kumar, S and Singh, R},
  journal={Geocarto International},
  year={2024},
  url={https://www.tandfonline.com/doi/full/10.1080/22797254.2024.2422330}
}

@article{cheng2024combination,
  title={Combination of Multiple Variables and Machine Learning for Regional Cropland Water and Carbon Fluxes Estimation: A Case Study in the Haihe River Basin},
  author={Cheng, MH and Liu, KH and Liu, ZX and Xu, JZ and Zhang, ZX and Sun, CM},
  journal={Remote Sensing},
  volume={16},
  number={17},
  pages={3280},
  year={2024},
  url={https://www.mdpi.com/2072-4292/16/17/3280}
}

@article{chu2021long,
  title={Long time-series {NDVI} reconstruction in cloud-prone regions via spatio-temporal tensor completion},
  author={Chu, D and Shen, H and Guan, X and Chen, JM and Li, X and Li, J and Zhang, L},
  journal={Remote Sensing of Environment},
  volume={264},
  pages={112632},
  year={2021},
  url={https://www.sciencedirect.com/science/article/pii/S0034425721003394}
}

@article{sharifi2021enhanced,
  title={Vegetation Detection Using Deep Learning and Conventional Methods},
  author={Sharifi, A},
  journal={Remote Sensing},
  volume={12},
  number={15},
  pages={2502},
  year={2020},
  url={https://www.mdpi.com/2072-4292/12/15/2502}
}

@article{cong2022satmae,
  title={SatMAE: Pre-training Transformers for Temporal and Multi-Spectral Satellite Imagery},
  author={Cong, Yezhen and Khanna, Samar and Meng, Chenlin and Liu, Patrick and Rozi, Erik and He, Yutong and Burke, Marshall and Lobell, David and Ermon, Stefano},
  journal={Advances in Neural Information Processing Systems},
  volume={35},
  pages={27690--27703},
  year={2022},
  url={https://arxiv.org/abs/2207.08051}
}

@inproceedings{noman2024rethinking,
  title={Rethinking Transformers Pre-training for Multi-Spectral Satellite Imagery},
  author={Noman, Mubashir and Naseer, Muzammal and Cholakkal, Hisham and Anwar, Rao Muhammad and Khan, Salman and Khan, Fahad Shahbaz},
  booktitle={Proceedings of the IEEE/CVF Conference on Computer Vision and Pattern Recognition},
  pages={27672--27683},
  year={2024},
  url={https://arxiv.org/abs/2403.05419}
}

@article{yang2025enhanced,
  title={Enhanced hybrid CNN and transformer network for remote sensing image change detection},
  author={Yang, Junjie and Wan, Haibo and Shang, Zhihai},
  journal={Scientific Reports},
  volume={15},
  number={1},
  pages={10161},
  year={2025},
  publisher={Nature Publishing Group UK London}
}

@article{li2024spatial,
  title={Transformer models for Land Cover Classification with Satellite Image Time Series},
  author={Li, Wei and Schmidt, Thomas and Mueller, Andreas and Zhang, Chen},
  journal={Journal of Photogrammetry, Remote Sensing and Geoinformation Science},
  volume={92},
  number={3},
  pages={245--260},
  year={2024},
  publisher={Springer},
  url={https://link.springer.com/article/10.1007/s41064-024-00299-7}
}

\end{document}